\documentclass[journal]{IEEEtran}
\usepackage{array}
\usepackage{indentfirst}
\usepackage{times}
\usepackage{epsfig}
\usepackage{graphicx}
\usepackage{amsmath}
\usepackage{amssymb}
\usepackage{amsmath,graphicx}
\usepackage{multirow} %//multirow for format of table
\usepackage{subfigure}
\usepackage{xcolor}
\usepackage{graphics}
\usepackage{epsf}
\usepackage{booktabs}
\usepackage{tabularx}
\usepackage{array,hhline}

\usepackage{amsmath,amsfonts,amssymb}
\usepackage{graphicx}

\usepackage{algorithm}
\usepackage{algorithmic}

\ifCLASSINFOpdf
\else
\fi
\begin{document}
\title{RGB Video Based Tennis Action Recognition Using a Deep Historical Long Short-Term Memory}

\author{Jiaxin~Cai,
        Xin~Tang,
        Haifeng~Xia,
        and~Meihong~Chi% <-this % stops a space
\thanks{Jiaxin Cai was with the School of Applied Mathematics, Xiamen University of Technology, Xiamen, Fujian, 361024 China,  e-mail: (caijiaxin@xmut.edu.cn ).}% <-this % stops a space
\thanks{Xin Tang was with the School of Electrical and Electronic Engineering, Nanyang Technological University,  e-mail: (tangxin@ntu.edu.sg).}% <-this % stops a space
\thanks{Haifeng Xia was with the XXX,  e-mail: (XXX ).}% <-this % stops a space
\thanks{Meihong Chi was with the XXX,  e-mail: (XXX ).}% <-this % stops a space
\thanks{Manuscript received April 19, 2005; revised August 26, 2015.}}

\maketitle

\begin{abstract}
Action recognition has attracted increasing attention from RGB input in computer vision partially
due to potential applications on somatic simulation and statistics of sport such as  tennis game. Recently, deep learning based methods have
achieved promising performance for action recognition. In this paper, we proposed a historical Long
Short-Term Memory adopted with convolutional neural network representations for three
dimensional tennis shots recognition. First, the local two-dimensional convolutional neural network
spatial representations were extracted from each video frame individually using a pre-trained
convolutional neural network. Then, a historical Long Short-Term Memory model was proposed to take the
output state at time t and the historical updated feature at time t-1 to generate a holistic feature vector using
a score weighting scheme. Finally, we used the adopted CNN and deep historical LSTM to map the
original visual sequences of tennis action into a  spatial-temporal semantical description
 and classified the action video content. Experiments on the benchmarks
demonstrate that our method which used only simple raw RGB video can achieve better performance than
the state-of-the-art baselines for tennis shot recognition.
\end{abstract}


% \fi
\IEEEpeerreviewmaketitle

\section{Introduction}

Due to the great increasing of sport game videos, it is implementable to analysis the technology and movement of players using the collected video data. Such as \cite{Bertasius} tries to evaluate a basketball player's performance from his/her first-person video. Many works have be devoted in vision-based sport action recognition in recent years \cite{Mora, Soomro}.
Several public video datasets such as  UCF-Sport \cite{Soomro} and Sports-1M \cite{Karpathy} are provided for this problem. Tennis has received widely attention from all the world.  However, since lack of data, there are few literatures analyzing tennis actions. This paper focuses on automatically recognizing different tennis actions using the RGB videos. Zhu et al. \cite{Zhu1,Zhu2} used the  support vector machine to classify tennis videos into left-swing and right-swing actions by optical flow based descriptor.
Farajidavar et al. \cite{Farajidavar}  employed the transfer learning to classify  tennis videos into non-hit, hit and serve actions.
However, they have just analyzed a few actions for playing tennis.
Luckily, Gourgari et al. \cite{Gourgari} have presented a tennis actions database called THETIS dataset.
It consists of 12 fine-grained tennis shots acted by 55 different subjects multiple times at different scenes with dynamic background. The objective of analyzing the THETIS dataset is  to classify the videos into the 12 pre-defined tennis actions from raw video data.
Based on this database, Mora et al. \cite{Mora} have proposed a deep learning model for domain-specific tennis action recognition using RGB video content.
They tried to use long-short term memory networks (LSTMs) to describe each action, since LSTMs can remember information for long periods of time.

Recently, deep learning based methods have achieved promising performance for action recognition. Li et al. \cite{Li} presented a skeleton-based action recognition method using LSTM and convolutional neural network (CNN).
Cheron et al. \cite{Cheron} proposed a pose-based convolutional neural network feature for action recognition.
%Gammulle et al. \cite{Gammulle} presented a two-stream LSTM framework to fuse the deep networks human action recognition.
Gammulle et al. \cite{Gammulle} have presented a deep fusion framework for human action recognition, which employs a convolutional neural network  to extract salient spatial features and LSTMs to model temporal relationship.
Zhang et al. \cite{Zhang} used the multi-layer LSTM networks to learn the geometric features on skeleton information for action recognition. Liu et al. \cite{Liu} proposed a spatio-temporal LSTM with trust gates for 3D human action recognition. Zhu et al. \cite{Zhu} presented a co-occurrence skeleton feature learning based on regularized deep LSTM networks  for human action recognition. Lee et al. \cite{Lee} proposed an ensemble learned temporal sliding LSTM networks for skeleton-based action recognition. Tsunoda et al. \cite{Tsunoda}   used the hierarchical LSTM model for football action recognition. Song et al. \cite{Song} presented a spatio-temporal attention-based LSTM networks for recognizing and detecting 3D action.
Liu \cite{Liu2} proposed a global context-aware attention LSTM networks for 3D action recognition.

At least two important aspects influence the performance of action recognition: spatial representation  and temporal modeling. Hence, in this paper, we propose a framework using convolutional neural networks with historical Long Short-Term Memory networks for tennis action recognition. First, we extract deep spatial representation for each frame by Inception \cite{szegedy2016rethinking} which is a famous convolutional neural network and pre-trained on ImageNet dataset. The most importance is that a historical Long Short-Term Memory units are proposed to modeling the temporal cues. 
For THETIS dataset, there are 12 fine-grained tennis shots and they are similar with each other. It's a challenge to distinguish different tennis shots. As we know, for each action, its historical evolution as time is important. 
Long Short-Term Memory is widely used for sequence model.
The original Long Short-Term Memory only model the local representation of sequences. However, the global or holistic representation, such as historical information can not be described through LSTM.
Therefore, we have proposed an recurrent neural network(RNN) architecture using an extra layer to describe the historical information for action recognition.
Besides, for a video clip, some frames may be very critical to action recognition. However, in original Long Short-Term Memory model, we can't automatically focus on those key frames. Therefore, we present a score weighting scheme by taking the output state at time $t$ and the historical information at time $t-1$ to generate the feature vector that describing the historical information of human action. In our framework, we can automatically pay more attention to those impressive frames and omit the insignificant clips.
Additionally, the update duration of historical state may be very long  so that large errors  may be  accumulated in the iteration process.
So we proposed a error truncation technology to   re-initialize the  historical state at current time and drop the accumulated errors.
In order to extract temporal features, we developed a five-layer deep  historical Long Short-Term Memory network to learn the representation of  the spatial feature sequences.
In conclusion, the CNN model and the deep historical LSTM model are adopted to map the
raw RGB video  into a vector space to generate the spatial-temporal semantical description
 and classify the action video content.
Experiments on the benchmarks demonstrate that our method outperforms the state-of-the-art baselines for tennis shot recognition using only  raw RGB video.

The main contributions of our work are listed as follows

(1) We proposed a Long Short-Term Memory units  based RNN architecture with an extra historical layer, called historical Long Short-Term Memory, to model the holistic information and build high level representation for visual sequences.

(2) In our proposed historical Long Short-Term Memory framework, a  historical layer was introduced to model the  historical information of hidden state.
So the proposed  model can learn the global and holistic feature from the original  human action sequence.

(3) The proposed historical Long Short-Term Memory used a weighting scheme to enhance the influence of key frames that are important to action  recognition during the updating process of historical state.
So it help the pattern classier to automatically pay more attention to the important frames and omit the insignificant clips.

(4)Due to the update duration of historical state is long  and large errors  may be  accumulated in the update process, we proposed a error truncation technology to re-initialize the   historical state at current time and drop the accumulated errors.

(5) We proposed a framework based on CNN and deep historical Long Short-Term Memory networks for tennis action recognition .
Firstly ,we used the pre-trained Inception networks to build the spatial features from raw RGB frames.
Then we proposed a five-layer historical Long Short-Term Memory networks for action video recognition.
The sequences of spatial feature vectors were fed into this deep historical Long Short-Term Memory model to build the temporal feature of videos.

The rest of the paper is organized as follows. Section 2 describes the proposed method. Section 3 shows the experimental results and analysis. Finally, section  gives the conclusion  and future work of  the paper.

\section{Method}

\subsection{Framework}

In this paper, we adopted an improved Long Short-Term Memory network accompanied by convolutional neural network  for  tennis action recognition.
The whole framework is illustrated in Figure 1. Firstly, each frame of the original RGB video is fed to the Inception V3 model for capturing the appearance and spatial information as the local feature. This model is pre-trained on the Large Scale Visual Recognition Challenge 2012 (ILSVRC-2012) ImageNet.
Then a  Long Short-Term Memory model is   adopted  as a complement to the CNN model to capture the contextual information in the temporal domain.
The spatial features extracted from Inception model are fed to the LSTM model as the input. The CNN features are suitable to feed into the LSTM model because they provides rich spatial information. Specially, we introduced an improved version of the Long Short-Term Memory, called historical Long Short-Term Memory, to describe the historical information as the global feature.
It employs a score weighting scheme  to generate the iterated feature vector using the output state at time $t$ and the historical state at time $t-1$.
Then a five-layer deep historical LSTM network is builded. Each layer of it has 30 historical LSTM units.
Finally, we use the output of deep historical LSTM networks as the spatial-temporal semantical description of visual sequences. It is fed to a soft-max layer for classifying the action video content.

\begin{figure}[htbp]
\centering
\includegraphics[height=12cm,width=8cm]{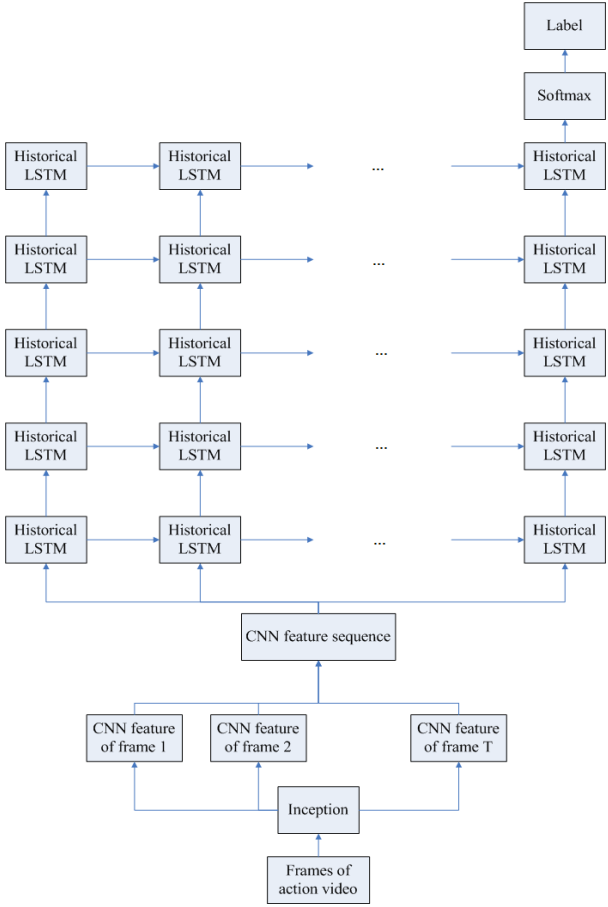}
\caption{The proposed framework for tennis action recognition}
\label{Fig1}
\end{figure}

\subsection{RNN and LSTM}

RNN is a powerful sequence  model  widely used in speech recognition and natural language processing.
For a typical RNN model, the update equation at time $t$ is described as follows.

\begin{equation} \label{eq1}
 \begin{aligned}
h_t=g(b+Wh_{t-1}+Ux_t)
\end{aligned}
\end{equation}

\begin{equation} \label{eq1}
 \begin{aligned}
\hat{y}_{t}=softmax(c+Vh_t)
\end{aligned}
\end{equation}

where $h_t$ is the state of the $t$-th  neuron, and $x_t$ is the observation at  time $t$ which is the input of the  $t$-th  neuron.
In a typical RNN model, the state response $h_t$ of the $t$-th neuron is determined by the previous neuron state $h_{t-1}$ and the input $x_t$ of the $t$-th neuron.
$g(\cdot)$ is the activation function.
Usually, the sigmoid function $ \sigma(\cdot) $ or hyperbolic tangent function $tanh(\cdot)$  can be employed as the activation function.
$\hat{y}_t$ denotes the prediction of label $y$ at  time $t$.
It is estimated by the output response $h_t$ of state neuron at $t$ time using a softmax function.
$U$, $W$ and $V$ are the weight matrices.
$b$ and $c$ are the bias vectors.

Traditional RNNs is suitable for  modeling the short-term dynamics but unable to capture the long-terms relations.
LSTM is an improved version of RNN architecture for learning long-range dependencies and resolving the "vanishing gradient" problem.
In the typical LSTM model, the $t$-th hidden neuron is the cell unit containing  an input gate $i_t$, a forget gate $f_t$,  output gate $o_t$,    an internal memory cell state $c_t$ and an output response $h_t$.
The transition equations of LSTM at  time $t$ can be written as follows.

\begin{equation} \label{eq1}
 \begin{aligned}
i_t=\sigma(U^{i}x_t+W^{i}h_{t-1}+P^{i}c_{t-1}+b^i)
\end{aligned}
\end{equation}

\begin{equation} \label{eq1}
 \begin{aligned}
f_t=\sigma(U^{f}x_t+W^{f}h_{t-1}+P^{f}c_{t-1}+b^f)
\end{aligned}
\end{equation}

\begin{equation} \label{eq1}
 \begin{aligned}
c_t=f_t \odot c_{t-1} +i_t \odot tanh(U^{c}x_t+W^{c}h_{t-1}+b^c)
\end{aligned}
\end{equation}

\begin{equation} \label{eq1}
 \begin{aligned}
o_t=\sigma(U^{o}x_t+W^{o}h_{t-1}+P^{o}c_{t}+b^o)
\end{aligned}
\end{equation}

\begin{equation} \label{eq1}
 \begin{aligned}
h_t=o_t \odot tanh(c_{t})
\end{aligned}
\end{equation}

where $U^{i}$, $U^{f}$, $U^{c}$, $U^{o}$, $W^{i}$, $W^{f}$, $W^{c}$, $W^{o}$, $P^{i}$, $P^{f}$ and $P^{o}$ are the weight matrices.
$b^i$, $b^f$, $b^c$ and $b^o$  are the bias vectors.
Operator $\odot$ indicates element-wise product.

\subsection{Historical  LSTM}

The typical LSTM use the last layer to encode the input sequence.
When it is adopted for action video recognition using  image sequences, the last frame of the video will have the most influence on the action class recognition result.
And the previous frames have a minimal impact on the classification due to the forgetting effect.
However, a human action is often defined by the whole movement process.
To better describe the human action, we embed the historical information of human movement to the LSTM model for building a global feature.
We propose an improved version of the Long Short-Term Memory network, called historical Long Short-Term Memory, to describe the historical information.
The structure of  model  is illustrated in Figure 2.
The historical Long Short-Term Memory network is  an LSTM units based RNN architecture using an added extra layer called historical layer to describe holistic information of human movement .
In the historical  Long Short-Term Memory model, a historical state $l_t$ is introduced at time $t$.
It is generated by a score weighting scheme  using the  response state $h_t$ at time t and the historical state  at time $t-1$.
If the classification loss of historical state at time $t-1$ is smaller than that of response state at time $t$,   then we generate the new historical state at time $t$ using the weighted sum of the current response state at time $t$ and  the historical state at time $t-1$.
The weight of the response state or the historical state is computed according to its classification loss.
Using such weighting scheme,  the influence of key frames which are important to action class recognition can be enhanced during the updating process of historical state.
So it help the pattern classier to automatically pay more attention to the important frames and omit the insignificant clips.
However, if the classification loss of historical state at time $t-1$ is lager than that of response state at time $t$, which means  too much errors have been accumulated in the long process of the historical state iteration,  then we recalculate the  current historical state at time $t$ using the response states from time $t-\tau$ to time $t$.
Here $\tau$ is a parameter to control the length of state sequence used to re-initialize the  historical state and drop the previous historical state with lager errors.
It determines when to forget the previous response states if they are corresponding to large errors.
The update equation of historical state $l_t$ is expressed as follows.

\begin{figure}[htbp]
\centering
\includegraphics[height=6cm,width=8cm]{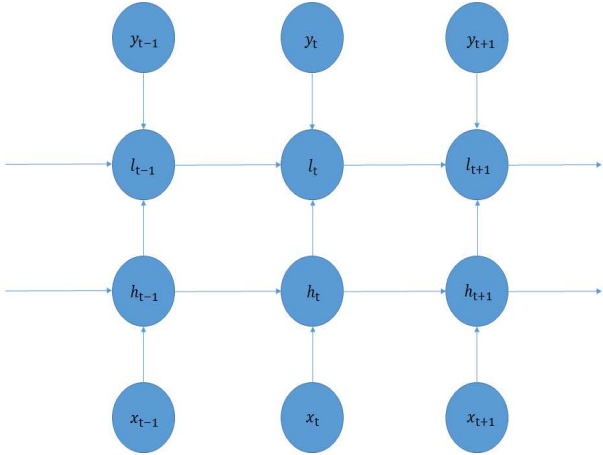}
\caption{The structure of  historical Long Short-Term Memory model}
\label{Fig1}
\end{figure}

\begin{equation} \label{eq1}
 \begin{aligned}
{l_t } = \left\{ \begin{array}{l}
\alpha_{t} h_{t} +(1-\alpha_{t}) l_{t-1} ,\ \ {\rm{ }} if \ \  {\rm{ }} \epsilon_{h_t} \geq \epsilon_{l_{t-1}}  \\
\sum_{k=1}^{t}\omega_{k}^{t}h_{k},\ \ {\rm{ }} if \ \  {\rm{ }} \epsilon_{h_t}<\epsilon_{l_{t-1}}
\end{array} \right.
\end{aligned}
\end{equation}

where $\alpha_t$ is the weight controlling a balance between the  response $h_t$ and the last historical state $l_{t-1}$.
It is calculated by the following formula:

\begin{eqnarray}
\label{linear_scheme}
\alpha_t=\frac{1}{2}\ln(\frac{\epsilon_{l_{t-1}}}{\epsilon_{h_t}})
\end{eqnarray}

where $\epsilon_{h_t}$ denotes the loss between the training label $y_t$ at time $t$ and the estimated label $\hat{y}_t$ at time $t$ which is produced by the softmax function on $c+Vh_t$ . In this paper, the training label $y_t$ at  time $t$ is set as the action label of the training video.

$\omega_{k}^{t}$ denotes the weight of response $h_k$.
It is calculated by:

\begin{equation} \label{eq1}
 \begin{aligned}
\omega_{k}^{t} = \left\{ \begin{array}{l}
0 ,\ \ {\rm{ }} if \ \  {\rm{ }} k \leq \tau    \\
\frac{1}{t-\tau},\ \ {\rm{ }} if \ \  {\rm{ }} k > \tau
\end{array} \right.
\end{aligned}
\end{equation}

where $\tau$ is the parameter controlling the forgetting effect.

Finally, the last historical state $l_T$ are used as the output of historical LSTM model. 
It describe the holistic information of human movement history and can be used for action video recognition.
For example, a softmax layer can be  employed to provide the estimated label $\hat{y}$ based on the last historical state $l_T$:

\begin{equation} \label{eq1}
 \begin{aligned}
\hat{y}=softmax(d+Ql_T)
\end{aligned}
\end{equation}

where $d$ and $Q$ is the bias vector and weight matrix of the softmax layer respectively.

\subsection{Implementation Details}

TensorFlow and Python are employed  as the deep learning platform.
An NVIDIA GTX1080Ti GPU is adopted to run the experiments.
We use a five-layer deep  historical LSTM in the stacked network.
Each layer  has 30 units.
The  probability of dropout is set as 0.5.
The initial learning rate is set as 0.001 on  historical LSTM model.
The learning rate is set by exponentially decaying with a base of 0.96 every 100 000 steps during the training process.
The regularization value of the  historical LSTM model is set as 0.004.
The batch size fed to the model is set as 32.
The Adam Optimizer is used to trained the network for  historical LSTM model.

\section{Experimental results}

\subsection{General action recognition}

\subsubsection{Dataset}

We use the HMDB51 dataset \cite{Kuehne} to show the effectiveness of the proposed historical LSTM for general  action recognition tasks.
The HMDB51 dataset is a common used dataset including 6849 videos.
It consists of 51 human action categories including facial actions and body movement.
The original RGB videos of  HMDB51 dataset are used to test the proposed framework.

\subsubsection{Results}
The five-fold cross-validation strategy is used to split the dataset to the training set and  test set.
We compared the proposed historical LSTM to the typical LSTM.
The parameter $\tau$ of historical LSTM is  fine tuned from 2 to 5.
Table 1 shows the accuracy values  of the methods.
As is shown in the table, the average accuracy in prediction of the historical LSTM model achieves a best accuracy $0.73$ when the parameter $\tau$ is set as 2.
Furthermore, the average accuracy  of the historical LSTM model is $0.62$ when the parameter $\tau$ is set as 3.
When the parameter $\tau$ is set as 4, the historical LSTM model achieves an accuracy of $0.63$.
However, when $\tau$ is set as 5, the accuracy of  historical LSTM decline to $0.62$.
It is seen that setting the parameter  $\tau$ to a medium number according to the time period in the WTSM model is better.
Meanwhile, the average accuracy of the LSTM model is $0.47$.
The experimental results shows that the proposed historical LSTM model outperforms  LSTM model.
We also compared the proposed model to other state-of-the-art models using RGB data, such as  Spatial stream ConvNet \cite{Simonyan}, Soft attention model \cite{Sharma}, Composite LSTM \cite{Soomro}  and Mora's model \cite{Mora}.
As shown in Table 1, the correct recognition rate of our method exceeds that of  the other models using RGB data.
We also compared our model using RGB data to other methods using different information.
They are  the two-stream ConvNet using optical flow information \cite{Simonyan},  Wang's model  using
Fisher Vectors \cite{Wang},  and Wang's model using  a combination of HOG, HOF
and MBH \cite{Wang}.
The experimental results  show that our model achieves a high accuracy using the very simple RGB data and outperform the compared  models.
It will results in better results if other information such as optical flow data, skeletons data and depth information  are added to the input of our proposed model.

\begin{table}[htbp]
  \centering
  \caption{Accuracy comparison  of methods on HMDB51 dataset}
    \begin{tabular}{cc}
    \toprule
    Method &  Accuracy \\
    \midrule
     Historical LSTM ($\tau=2$)  & 0.73\\
     Historical LSTM ($\tau=3$)  & 0.62 \\
     Historical LSTM ($\tau=4$)  & 0.63 \\
     Historical LSTM ($\tau=5$)  & 0.62 \\
    LSTM & 0.47 \\
    Spatial stream ConvNet \cite{Simonyan} & 0.41 \\
    Soft attention model \cite{Sharma} & 0.41 \\
    Composite LSTM \cite{Soomro} & 0.44 \\
    Mora et al. \cite{Mora} & 0.43 \\
    Two-stream ConvNet using optical flow  \cite{Simonyan}  & 0.59 \\
    Wang et al. (using Fisher Vectors)  \cite{Wang} & 0.53 \\
    Wang et al. (using  a combination of HOG, HOF and MBH)  \cite{Wang} & 0.60 \\
    \bottomrule
    \end{tabular}%
  \label{tab:recresult}%
\end{table}%

\subsection{Tennis action recognition}

\subsubsection{Dataset}

The Three Dimensional Tennis Shots (THETIS) dataset  \cite{Gourgari} is used for evaluating the proposed method.
This  dataset contains  12 basic tennis actions each of which are acted repeatedly by 31 amateurs and 24 experienced players.
The 12 tennis actions performed by actors are: Backhand with two hands,  Backhand, Backhand slice, Backhand volley,  Forehand flat,  Forehand open stands,  Forehand slice,  Forehand volley, Service flat,  Service kick,  Service slice and  Smash.
Each actor repeats each tennis action 3 to 4 times.
There are totally 8734 videos of the AVI format  with a total duration of  7 hours and 15 minutes.
Specially, a set of 1980 RGB viodeos of the AVI format are provided.
There are two different indoor backgrounds.
The backgrounds contain  different scenes in which multiple persons pass or play basketball.
The length of video sequences also varies.
Although the depth, skeleton  and silhouettes videos are also given,  we only use the RGB videos to perform the tennis action recognition experiments.

\subsubsection{Results}
The five-fold cross-validation strategy is used to split the dataset to the training set and test set.
We compared the proposed historical LSTM to the typical LSTM.
The parameter $\tau$ of historical LSTM is  fine tuned from 2 to 5.
Table 2 shows the recognition accuracy values  of the methods.
As is shown in the table, when the parameter $\tau$ is set as 3, the average accuracy of the historical LSTM model achieves an accuracy of $0.74$.
Furthermore, the average accuracy  of the historical LSTM model is $0.70$ when the parameter $\tau$ is set as 2.
When the parameter $\tau$ is set as 4, the historical LSTM model achieves an accuracy of $0.71$.
However, when $\tau$ is set as 5, the accuracy of  historical LSTM decline to 0.63.
It is seen that setting the parameter  $\tau$ to a medium number according to the time period in the historical LSTM model is better.
Meanwhile, the average accuracy of the LSTM model is $0.56$.
The experimental results shows that the proposed historical LSTM model outperforms the LSTM model in the accuracy of perdition.

\begin{table}[htbp]
  \centering
  \caption{Accuracy comparison of  methods on THETIS dataset}
    \begin{tabular}{cc}
    \toprule
    Method &  Accuracy \\
    \midrule
     Historical LSTM ($\tau=2$)  & 0.70 \\
     Historical LSTM ($\tau=3$)  & 0.74 \\
     Historical LSTM ($\tau=4$)  & 0.71 \\
     Historical LSTM ($\tau=5$)  & 0.63 \\
    LSTM & 0.56 \\
    Mora et al. \cite{Mora} & 0.47 \\
   Gourgari et al. (using depth videos)\cite{Gourgari} & 0.6 \\
   Gourgari et al. (using 3D skeletons)\cite{Gourgari} & 0.54 \\
    \bottomrule
    \end{tabular}%
  \label{tab:recresult}%
\end{table}%

In \cite{Mora}, the authors performed experiments using the RGB videos on the THETIS dataset  and get an average accuracy of 0.47 for tennis action recognition using a leave-one-out strategy.
Compared to their experiments, we employ less training samples and obtain a better performance under a more difficult condition.
In \cite{Gourgari} , the authors performed tennis action classification experiments  using depth videos and 3D skeletons separately on the THETIS dataset and obtained accuracy 0.60 and 0.54 respectively.
Compared to their results, our experiment use the raw images alone, which is a more challenge task.
Experimental results prove that our proposed method is competitive and outperform the-state-of-art methods.

\section{Conclusion}

This paper  proposed a  historical Long Short-Term Memory  for modeling the tennis action image sequence.
A score weighting strategy was introduced to the Long Short-Term Memory networks to model the historical information.
The proposed historical LSTM was used for tennis action recaption on the CNN features of local frame images of videos.
Experimental results on the tennis action datasets  demonstrate that our method is effective.
Our future works will include the improving of the weighting strategy and more experiments on real tennis game videos.

% or


\section*{Acknowledgment}

This work was supported by  the Xiamen University of Technology High Level
Talents Project (No.YKJ15018R).

\ifCLASSOPTIONcaptionsoff
  \newpage
\fi

\end{document}